\documentclass[11pt,a4paper]{article}
\usepackage[hyperref]{acl2021}
\usepackage{times}
\usepackage{inconsolata}
\usepackage{multirow}
\usepackage{latexsym}
\usepackage{listings}

\usepackage[normalem]{ulem}
\usepackage{microtype}
\usepackage{tikz}
\def\checkmark{\tikz\fill[scale=0.4](0,.35) -- (.25,0) -- (1,.7) -- (.25,.15) -- cycle;}

\usepackage{subcaption}

\usepackage{xcolor}

\definecolor{codegreen}{rgb}{0,0.6,0}
\definecolor{codegray}{rgb}{0.5,0.5,0.5}
\definecolor{codepurple}{rgb}{0.58,0,0.82}
\definecolor{backcolour}{rgb}{0.95,0.95,0.92}

\lstdefinestyle{mystyle}{
    backgroundcolor=\color{backcolour},   
    commentstyle=\color{codegreen},
    keywordstyle=\color{magenta},
    numberstyle=\tiny\color{codegray},
    stringstyle=\color{codepurple},
    basicstyle=\ttfamily\footnotesize,
    breakatwhitespace=false,         
    breaklines=true,                 
    captionpos=b,                    
    keepspaces=true,                 
    showspaces=false,                
    showstringspaces=false,
    showtabs=false,                  
    tabsize=2
}

\usepackage{multirow}
\usepackage{booktabs}

\aclfinalcopy 
\def\aclpaperid{99} 

\usepackage[normalem]{ulem}

\def\ODdel#1{\bgroup\markoverwith{\textcolor{purple!60}{\rule[0.4ex]{2pt}{3pt}}}\ULon{#1}}

\title{Leveraging Large Language Models for Building Interpretable Rule-Based Data-to-Text Systems}

\author{J\c edrzej Warczyński$^1$ \and Mateusz Lango$^{1,2}$ \and Ond\v rej Du\v sek$^2$ \\
  $^1$Poznan University of Technology,  Faculty of Computing and Telecommunications, Poznan, Poland  \\
  $^2$Charles University, Faculty of Mathematics and Physics, Prague, Czechia \\
  \texttt{jedrzej.warczynski@student.put.edu.pl},   \texttt{\{lango,odusek\}@ufal.mff.cuni.cz} \\}

\date{}

\begin{document}
\maketitle
\begin{abstract}
We introduce a simple approach that uses a large language model (LLM) to automatically implement a fully interpretable rule-based data-to-text system in pure Python. 
Experimental evaluation on the WebNLG dataset showed that such a constructed system produces text of better quality (according to the BLEU and BLEURT metrics) than the same LLM prompted to directly produce outputs, and produces fewer hallucinations than a BART language model fine-tuned on the same data.
Furthermore, at runtime, the approach generates text in a fraction of the processing time required by neural approaches, using only a single CPU.
\end{abstract}

\section{Introduction}
Data-to-text is a field of natural language generation (NLG) that focuses on converting structured, non-linguistic data into coherent text~\cite{gattkrahmer}. This paper, like many others in the field~\cite{castro-ferreira-etal-2020-2020,agarwal-etal-2021-knowledge,kasner-dusek-2022-neural}, specifically addresses the challenge of generating text from data expressed as RDF triples that consist of a subject, a predicate, and an object. For instance, one possible textualization of the following RDF triples: (Mozart, birthplace, Vienna), (Mozart, birth year, 1756) is ``Mozart was born in 1756 in Vienna.''

There are two main approaches to the construction of data-to-text systems: rule-based and neural methods~\cite{gattkrahmer}.
Rule-based approaches~\cite{lavoie-rainbow-1997-fast,white-baldridge-2003-adapting} rely on predefined templates and linguistic rules to transform structured data into text, ensuring high precision and control over the output. 
On the other hand, neural approaches leverage deep learning models to automatically learn the mapping from data to text~\cite{ke-etal-2021-jointgt,chen-etal-2020-kgpt}. They offer greater flexibility and produce more natural and varied text, but have limited interpretability, are more computationally intensive and prone to producing hallucinations~\cite{rebuffel_controlling_2022,10.1145/3571730}.

This paper combines these two perspectives on building NLG systems and proposes to use a large neural language model to \emph{train} (implement) a rule-based system.
Specifically, we propose a training procedure that processes the training set by asking a large language model to write simple Python code that would generate the reference text based on the input data.
The generated code is executed to check for syntax errors and whether it produces the correct output. 
The final result of the training of the system is a single file of Python code that is able to generate a textualisation for the input data.

Although experimental evaluation on the WebNLG dataset~\cite{gardent-etal-2017-creating} showed that our automatically written rule-based system does not achieve the performance of a fully fine-tuned neural model in terms of BLEU or BLEURT score, it produces significantly fewer hallucinations and outperforms a non-trivial neural baseline on these measures.
Moreover, our system is fully interpretable and offers high controllability, as it can be modified by a Python programmer if necessary. Our approach also does not require a GPU during inference and produces text almost instantaneously on a single CPU. 





\begin{figure*}[t]
    \centering
    \includegraphics[width=\linewidth]{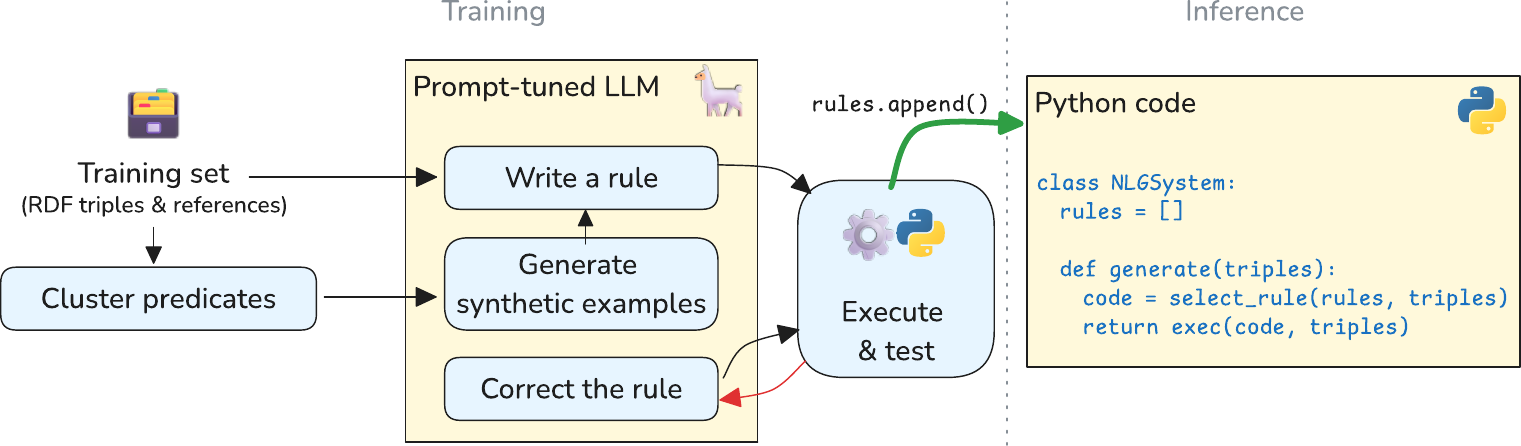}
    \caption{An overview of the training process of our rule-based system. Note that the output of the training process is a NLG system implemented in pure Python code that does not need access to the LLM to generate text. }
    \label{fig:enter-label}
\end{figure*}

\section{Target rule-based system structure}
\label{sec:structure}

We conceptualize a high-level fixed structure for our proposed system's Python code which
organises processing according to the set of predicates present in the input triples.
It contains two main elements: (1) an (initially empty) \emph{list of rules} capable of converting a set of triples with particular predicates into text, and (2) a \emph{rule selector} that processes the input triples and executes the corresponding rules.

Each rule is a plain Python code snippet/subroutine, coupled with with simplistic specifications of the expected input, including the expected number of triples and the list of their expected predicates. The rules are arranged in a simple list.
Before a rule's code is executed, the input triples are always sorted to match the order of the predicates given in the rule's specification. This allows simpler rules to be written and limits the number of potential errors. 

The rule selector processes the input triples by extracting their predicates and executing the rule that has the same list of predicates in the specification.
If there is no matching rule, the input is split into several parts by a splitting mechanism that aims to minimize the number of splits by applying greedy search. 
It iteratively searches for a rule capable of processing the largest subset of input triples, executes it, eliminates the already processed triples from the input and repeats the process.
If no rule can be found by further splitting, the triples are converted to text by a default rule ``\{subject\} \{predicate\} \{object\}".

\section{Training: LLM-based rule generation}
The goal of the training procedure is to populate the list of rules with useful rules capable of producing a fluent and hallucination-free description of the input triples. 

First, the approach makes a single pass through the training set, writing for each training example a Python code capable of producing the reference text (Sec.~\ref{sec:genrul}).  The training procedure only analyses instances that are not fully covered by already trained rules (i.e.~they cannot be processed without applying the splitting mechanism), which significantly reduces the size of the training set effectively needed to train the system.

Next, the approach uses a simple mechanism to improve the generalisability of the constructed system (Sec.~\ref{sec:generalizability}). The triples from the training set are clustered to discover sets of predicates that are likely to occur together on the input. Then, for each likely set of predicates, an artificial training example is constructed by interacting with an LLM, and then a standard rule construction procedure is applied.

\subsection{Generating rules from training examples}\label{sec:genrul}
The procedure for constructing a single rule for a given training instance consists of the three following steps:
    
\paragraph{Step 1: Prompt the LLM to write a rule}
    The LLM is instructed to generate Python code that produces a factual textual description of the data given in the input. Both the triples and the expected output (reference text) are provided in the prompt, but the model is informed that the code should be general enough to produce correct text even if the subjects/objects given in the triples are changed. A simple code snippet is also included in the prompt to inform the model about the classes used to represent the input and the general structure of the code. See the full prompt in Appendix~A.

\paragraph{Step 2: Execute and test the rule}
    The code of the rule is extracted from the response provided by the LLM, and simple formatting heuristics are applied to correct minor issues such as incorrect code indentation. The code is then executed in a separate process with a predefined timeout. If the code terminates before the timeout, does not throw an error, and the Levenshtein distance between the output text and the reference is within a predefined range, the rule is considered correct and added to the list of rules. Otherwise, the rule is regarded as incorrect.

\paragraph{Step 3: Correct the rule if needed}
    If the rule written by the LLM is incorrect, the model is informed about the incorrect output produced or the error returned, and it is asked to correct the issue (see the prompt in Appendix~\ref{app:prompts}). This process is repeated twice. If the returned code is still incorrect, the generation process is restarted from scratch, beginning a new conversation with the model to write the rule (Step 1). If this procedure fails a second time, rule construction is skipped for the given training instance.

\subsection{Generating additional rules for improved generalization}\label{sec:generalizability}

As mentioned above,  we generate additional rules for predicates that are likely to occur together in a sentence to improve the generalisation of the constructed rule-based system.

\paragraph{Clustering predicates}
To cluster predicates from the training set, we have developed a simple graph clustering algorithm. We start by constructing a graph, where each node represents a predicate in the training set. We then add connections between nodes (predicates) that co-occur in at least one training instance. Each connected component in such a constructed graph represents an initial cluster of predicates.

Since some clusters are too large for further processing, we split connected components with more than 20 nodes by systematically removing nodes connected to all other nodes within the component.
After adjusting the cluster sizes, we generate training instances for all pairs, triples and quadruples of predicates belonging to the same cluster using the procedure described below.

\paragraph{Generating synthetic training examples}
To create a training instance for a given list of predicates, we again prompt the LLM.
The prompt includes an instruction to generate a full list of triples using the specified predicates (i.e., come up with some relevant subjects and objects for the predicates), along with a corresponding reference text. 
Several input-output examples from the training set are provided to the LLM for context. The number of these training examples varies to ensure coverage of all requested predicate textualisations. Specifically, we used the splitting procedure from the rule selector (see Sec.~\ref{sec:structure}) to divide the list of predicates, and then identified the relevant training examples for each part.
The template for the corresponding LLM prompt can be found in Appendix~\ref{app:prompts}.

\section{Experimental evaluation}
\begin{table*}[t]
\small
\centering
\begin{tabular}{l|ccc|cc|c}
\toprule
&&&& \multicolumn{2}{c|}{\textbf{inference time}} &\\
& \bf BLEU & \bf METEOR & \textbf{BLEURT} & \textbf{GPU} & \textbf{CPU} & \textbf{interpretability} \\\midrule
Prompted Llama 3 70B        & 38.26          & \uline{0.680}  & 0.113          & 6,360 s           & n/a                                  & $\times$   \\
Fine-tuned BART             & \textbf{53.28} & \textbf{0.716} & \textbf{0.257} & \phantom{0,}249 s & 1,910 s                   & $\times$   \\
Our rule-based approach (with Llama 3 70B)  & \uline{42.51}  & {0.671}        & \uline{0.157}  & -               & \phantom{0,00}\textbf{3 s} & \checkmark \\
\bottomrule                 
\end{tabular}
\caption{Results of automatic evaluation on the WebNLG test set using BLEU, METEOR and BLEURT. 
Additionally, the inference time (in seconds) for the full test set is reported. The reported times do not include loading the models into memory and were measured on a machine with an Nvidia A40 48 GB GPU and an AMD EPYC 7313 CPU.
}
\label{tab:opendial-gready}
\end{table*}

\begin{table*}[t]
\small\centering
\begin{tabular}{l|ccccc}
\toprule
& \multicolumn{2}{c}{\bf hallucinations} &  \\
\bf  & \bf minor & \bf major & \bf omissions & \bf disfluencies & \bf repetitions  \\
\midrule

Prompted Llama 3 70B & \uline{ 0.08}          & \textbf{0.07}         & \textbf{0.07} & 0.19          & \textbf{0.03}  \\
Fine-tuned BART &  0.20          & {0.33} & 0.19          & \uline{0.16} & 0.07         \\
Our rule-based approach (with Llama 3 70B) & \textbf{0.04}          & \uline{0.13}          & \uline{0.08}          & \textbf{0.13}          & \textbf{0.03}                 \\
\bottomrule
\end{tabular}
\caption{Results of manual evaluation on a sample of 75 examples from the WebNLG test set (percentage of examples with different types of errors, see Sec.~\ref{sec:human-eval} for details).}
\label{tab:manual}
\end{table*}

\subsection{Experimental setup}
\paragraph{Dataset}
We performed experiments on the WebNLG benchmark~\cite{gardent-etal-2017-creating} containing data expressed as RDF triples and corresponding text references, which is prominent in many previous works. The rule-based system was trained only on the training part of the dataset, the fine-tuned baseline additionally used the development part as a validation set. All systems were tested on the in-domain part of the test set.

\paragraph{Baselines}
We compare the results of our rule-based approach with two baselines:
\begin{itemize}
\item The BART-base model~\citep{lewis-etal-2020-bart} fine-tuned on WebNLG dataset with the default architecture for conditional language modelling provided by HuggingFace library~\cite{wolf-etal-2020-transformers}.
More training details are in Appendix~\ref{app:trainingdetails}.
\item A prompted LLM -- to generate textual descriptions for provided triples, we use the instruction-tuned 70B version of the Llama~3 model~\cite{touvron2023llama,llama3_2024}, in a quantized version through the \emph{ollama} library.\footnote{\url{https://ollama.com/}, model ID \texttt{llama3:70b}.}
A simple post-processing of the results was applied to remove superfluous text, such as encouragements for further interaction with the model.  
The prompt used is provided in Appendix~\ref{app:prompts}.
\end{itemize}

\paragraph{Our rule-based approach}
We run our procedure of training a rule-based approach with Llama 3 70B  large language model. The threshold of 5 on the Levenshtein distance is used to verify the correctness of a rule during training (see Sec.~\ref{sec:genrul}, step~2).
Training was performed on two NVidia L40 48GB GPUs with quantized models (FP8). The processing of the original WebNLG dataset took less than 7 hours (6h 56m) and resulted in the construction of 3,408 rules. The generation of additional rules (Sec.~\ref{sec:generalizability}) resulted in approximately 110k new rules.

\subsection{Automatic evaluation}
We investigate the quality of generated output using several popular metrics: BLEU~\cite{Papineni02bleu:a}, METEOR~\cite{meteor} 
and BLEURT~\cite{bleurt}. 
Implementations of these metrics from HuggingFace~\cite{wolf-etal-2020-transformers} are used.
The results are presented in Table~\ref{tab:opendial-gready}.

In terms of automatic text quality metrics, the fine-tuned BART model achieved the highest scores. However, our rule-based approach ranked second in both the BLEU and BLEURT metrics, outperforming the prompted Llama 3 model. 
Moreover, this result was computed on a single CPU 83 times faster than the fastest neural approach (BART) running on a GPU.
We also assessed the effect of the additional rules generated from synthetic data by evaluated a variant of the system without these rules. We found the effect on metrics to be minimal (BLEU gain of 0.3\%, BLEURT and METEOR stay within 0.001). Nevertheless, we still retain these rules to increase fluency for predicate combinations unseen in training data.

\paragraph{Experiments with different LLMs}
To investigate the impact of a particular selection of large language model, we additionally performed experiments with two smaller, general-purpose LLMs: Mistral 7B \cite{jiang_mistral_2023}, Llama~3 7B \cite{llama3_2024}, as well as with one model specially tailored for programming: Code Llama 7B \cite{roziere_code_2024}.\footnote{Corresponding ollama model IDs: \texttt{mistral}, \texttt{llama3}, \texttt{codellama:7b-instruct}.}
The results of automatic evaluation are presented in Table~\ref{tab:othermodels}.
It can be observed that the task of writing NLG rules is quite challenging for the language models, as there is a significant performance gap, especially in terms of BLEU, between the results of Llama~3 70B and smaller models.

\begin{table}[t]
\centering\small
\begin{tabular}{lrr}
\toprule
             & \bf BLEU          & \bf METEOR         \\\midrule
Llama 3 70B  & \textbf{42.51} & \textbf{0.671} \\
Llama 3 7B   & 39.70          & 0.670           \\
Mistral 7B   & 35.36         & 0.636          \\
Codellama 7B & 36.67         & 0.611         \\
\bottomrule
\end{tabular}
\caption{Results of automatic evaluation of our rule generation approach using different LLMs on the WebNLG test set using BLEU and METEOR metrics.}
\label{tab:othermodels}
\end{table}


\subsection{Human evaluation}
\label{sec:human-eval}

To validate the results obtained from automatic metrics, we conducted a small-scale in-house human evaluation. We selected 75 instances from the test set of the WebNLG dataset and evaluated the outputs of our approach and both baselines, totalling 225 system outputs.
Following our previous research~\cite{lango-dusek-2023-critic}, the annotation was performed by asking binary questions related to the existence of minor hallucinations (such as typos in named entity names), major hallucinations (output containing facts not supported by the data), omissions (missing information), disfluencies (grammar errors or difficult-to-read text), and repetitions (information mentioned twice). 
The annotation was performed by five NLP experts, each output was evaluated by a single annotator.
The annotators were shown the input triples along with corresponding outputs from all three evaluated systems.
The annotation process was blinded, with the system outputs order randomly shuffled for each example.

\paragraph{Results}
The results are presented in Table~\ref{tab:manual}.
The proposed rule-based approach produces fewer minor hallucinations than both neural counterparts, has the lowest number of disfluencies and, ex aequo with the prompted LLM, the lowest number of repetitions.
The model also makes omissions at a frequency comparable to prompted LLM and significantly lower than fine-tuned BART.
In terms of major hallucinations, the proposed approach offers a statistically significant improvement over fine-tuned BART\footnote{Confirmed by a two-sample T-test for proportions with continuity correction, with $p=0.006$.}, but falls short of the prompted LLM.
We hypothesise that the gap between our system and LLM is a result of error accumulation: our system is partially trained with silver-standard, LLM-generated references that may contain hallucinations, and also suffers from potential errors in the written rules.
There  is also a possibility that the LLM results on generating outputs from  WebNLG dataset are affected by data leakage~\cite{balloccu-etal-2024-leak}, which is not the case for generating rules that are not present in the original dataset.

\paragraph{Human intervention experiment}
Since the manual evaluation identified several hallucinations produced by a rule-based system, we assessed the human effort required to fix them.
We randomly selected five examples with hallucinations and asked an experienced Python programmer to fix the code.
The programmer was able to use a standard IDE, but without the support of AI tools such as Copilot. 
The average time to fix one example was three minutes. In the automatic evaluation performed, none of the automatic metrics showed any degradation in the quality of the results, and the results for all selected examples were correct.
This demonstrates the interpretability and controllability of the generated rule-based system.

\paragraph{How do the rules looks like?}
The code of a typical rule has 5 lines of code (median) and very often contains renaming or extracting data from the input into a custom data structure (e.g. a dictionary, defaultdict, list) and then filling a textual template.
The final text is often constructed by iterating over the input triples or custom data structure and appending parts of the sentence to the output.
However, some of the rules are quite complex as they list possible conversions of data into text according to the context (e.g. a list how to convert month number into a month name).
The code of the longest rule produced has 51 lines.
Several examples of written rules are provided in Appendix~\ref{app:examplesrules}.

\section{Summary}
We presented a new way of training NLG systems for data-to-text problems: we use a large black-box language model to write fully interpretable Python code that is able to generate data textualisation in a fraction of the processing time required by fully neural systems. 
The experimental evaluation showed that the quality of the generated text is somewhere between that of a few-shot prompted LLM and BART finetuned on the same training data, offering an interesting trade-off between computational and training data requirements, interpretability and predictive performance.
In future work, we will extend the synthetic data generation to out-of-domain situations. We also plan to include new types of rules, such as rules operating at the sentence level (e.g. adding subordinate clauses).

\section*{Limitations}
Currently, our approach does not allow the generation of rules for unseen, i.e. out-of-domain predicates. This could be circumvented by providing a list of out-of-domain relations or even examples of out-of-domain inputs (without reference texts) to our clustering mechanism (Sec.~\ref{sec:generalizability}). Alternatively, these procedures could be applied on-the-fly, but this would require access to an LLM during inference.

The presented approach may also generate hallucinated (i.e.~non-factual) outputs, but the experiments demonstrated that the number of hallucinations is smaller than in the text generated by a fine-tuned transformer-based language model.

\section*{Supplementary Materials Availability Statement} 
Source code is available in our GitHub repository.\footnote{\url{https://github.com/jwarczynski/RuLLeM}}
All experiments were performed on the version of WebNLG dataset available through the HuggingFace Hub.\footnote{\url{https://huggingface.co/datasets/webnlg-challenge/web_nlg}}

\section*{Acknowledgments} Co-funded by the European Union (ERC, NG-NLG, 101039303) and National Science Centre, Poland (Grant No.~2022/47/D/ST6/01770).
This work used resources of the LINDAT/\hspace{0mm}CLARIAH-CZ Research Infrastructure (Czech Ministry of Education, Youth, and Sports project No.~LM2018101).
For the purpose of Open Access, the author has applied a CC-BY public copyright licence to any Author Accepted Manuscript (AAM) version arising from this submission.

\bibliographystyle{acl_natbib}
\bibliography{anthology,acl2021,custom}

\begin{thebibliography}{22}
\expandafter\ifx\csname natexlab\endcsname\relax\def\natexlab#1{#1}\fi

\bibitem[{Agarwal et~al.(2021)Agarwal, Ge, Shakeri, and
  Al-Rfou}]{agarwal-etal-2021-knowledge}
Oshin Agarwal, Heming Ge, Siamak Shakeri, and Rami Al-Rfou. 2021.
\newblock \href {https://doi.org/10.18653/v1/2021.naacl-main.278} {Knowledge
  graph based synthetic corpus generation for knowledge-enhanced language model
  pre-training}.
\newblock In \emph{Proceedings of the 2021 Conference of the North American
  Chapter of the Association for Computational Linguistics: Human Language
  Technologies}, pages 3554--3565, Online. Association for Computational
  Linguistics.

\bibitem[{Balloccu et~al.(2024)Balloccu, Schmidtov{\'a}, Lango, and
  Dusek}]{balloccu-etal-2024-leak}
Simone Balloccu, Patr{\'\i}cia Schmidtov{\'a}, Mateusz Lango, and Ondrej Dusek.
  2024.
\newblock \href {https://aclanthology.org/2024.eacl-long.5} {Leak, cheat,
  repeat: Data contamination and evaluation malpractices in closed-source
  {LLM}s}.
\newblock In \emph{Proceedings of the 18th Conference of the European Chapter
  of the Association for Computational Linguistics (Volume 1: Long Papers)},
  pages 67--93, St. Julian{'}s, Malta. Association for Computational
  Linguistics.

\bibitem[{Banerjee and Lavie(2005)}]{meteor}
Satanjeev Banerjee and Alon Lavie. 2005.
\newblock \href {https://www.aclweb.org/anthology/W05-0909} {{METEOR}: An
  automatic metric for {MT} evaluation with improved correlation with human
  judgments}.
\newblock In \emph{Proceedings of the {ACL} Workshop on Intrinsic and Extrinsic
  Evaluation Measures for Machine Translation and/or Summarization}, pages
  65--72, Ann Arbor, Michigan. Association for Computational Linguistics.

\bibitem[{Castro~Ferreira et~al.(2020)Castro~Ferreira, Gardent, Ilinykh,
  van~der Lee, Mille, Moussallem, and
  Shimorina}]{castro-ferreira-etal-2020-2020}
Thiago Castro~Ferreira, Claire Gardent, Nikolai Ilinykh, Chris van~der Lee,
  Simon Mille, Diego Moussallem, and Anastasia Shimorina. 2020.
\newblock \href {https://aclanthology.org/2020.webnlg-1.7} {The 2020 bilingual,
  bi-directional {W}eb{NLG}+ shared task: Overview and evaluation results
  ({W}eb{NLG}+ 2020)}.
\newblock In \emph{Proceedings of the 3rd International Workshop on Natural
  Language Generation from the Semantic Web (WebNLG+)}, pages 55--76, Dublin,
  Ireland (Virtual). Association for Computational Linguistics.

\bibitem[{Chen et~al.(2020)Chen, Su, Yan, and Wang}]{chen-etal-2020-kgpt}
Wenhu Chen, Yu~Su, Xifeng Yan, and William~Yang Wang. 2020.
\newblock \href {https://doi.org/10.18653/v1/2020.emnlp-main.697} {{KGPT}:
  Knowledge-grounded pre-training for data-to-text generation}.
\newblock In \emph{Proceedings of the 2020 Conference on Empirical Methods in
  Natural Language Processing (EMNLP)}, pages 8635--8648, Online. Association
  for Computational Linguistics.

\bibitem[{Gardent et~al.(2017)Gardent, Shimorina, Narayan, and
  Perez-Beltrachini}]{gardent-etal-2017-creating}
Claire Gardent, Anastasia Shimorina, Shashi Narayan, and Laura
  Perez-Beltrachini. 2017.
\newblock \href {https://doi.org/10.18653/v1/P17-1017} {Creating training
  corpora for {NLG} micro-planners}.
\newblock In \emph{Proceedings of the 55th Annual Meeting of the Association
  for Computational Linguistics (Volume 1: Long Papers)}, pages 179--188,
  Vancouver, Canada. Association for Computational Linguistics.

\bibitem[{Gatt and Krahmer(2018)}]{gattkrahmer}
Albert Gatt and Emiel Krahmer. 2018.
\newblock \href {https://arxiv.org/abs/1703.09902} {Survey of the state of the
  art in natural language generation: core tasks, applications and evaluation}.
\newblock \emph{J. Artif. Int. Res.}, 61(1):65–170.

\bibitem[{Ji et~al.(2023)Ji, Lee, Frieske, Yu, Su, Xu, Ishii, Bang, Madotto,
  and Fung}]{10.1145/3571730}
Ziwei Ji, Nayeon Lee, Rita Frieske, Tiezheng Yu, Dan Su, Yan Xu, Etsuko Ishii,
  Ye~Jin Bang, Andrea Madotto, and Pascale Fung. 2023.
\newblock \href {https://doi.org/10.1145/3571730} {Survey of hallucination in
  natural language generation}.
\newblock \emph{ACM Comput. Surv.}, 55(12).

\bibitem[{Jiang et~al.(2023)Jiang, Sablayrolles, Mensch, Bamford, Chaplot,
  Casas, Bressand, Lengyel, Lample, Saulnier, Lavaud, Lachaux, Stock, Scao,
  Lavril, Wang, Lacroix, and Sayed}]{jiang_mistral_2023}
Albert~Q. Jiang, Alexandre Sablayrolles, Arthur Mensch, Chris Bamford,
  Devendra~Singh Chaplot, Diego de~las Casas, Florian Bressand, Gianna Lengyel,
  Guillaume Lample, Lucile Saulnier, Lélio~Renard Lavaud, Marie-Anne Lachaux,
  Pierre Stock, Teven~Le Scao, Thibaut Lavril, Thomas Wang, Timothée Lacroix,
  and William~El Sayed. 2023.
\newblock \href {https://doi.org/10.48550/arXiv.2310.06825} {Mistral {7B}}.
\newblock ArXiv:2310.06825 [cs].

\bibitem[{Kasner and Dusek(2022)}]{kasner-dusek-2022-neural}
Zden{\v{e}}k Kasner and Ondrej Dusek. 2022.
\newblock \href {https://doi.org/10.18653/v1/2022.acl-long.271} {Neural
  pipeline for zero-shot data-to-text generation}.
\newblock In \emph{Proceedings of the 60th Annual Meeting of the Association
  for Computational Linguistics (Volume 1: Long Papers)}, pages 3914--3932,
  Dublin, Ireland. Association for Computational Linguistics.

\bibitem[{Ke et~al.(2021)Ke, Ji, Ran, Cui, Wang, Song, Zhu, and
  Huang}]{ke-etal-2021-jointgt}
Pei Ke, Haozhe Ji, Yu~Ran, Xin Cui, Liwei Wang, Linfeng Song, Xiaoyan Zhu, and
  Minlie Huang. 2021.
\newblock \href {https://doi.org/10.18653/v1/2021.findings-acl.223}
  {{J}oint{GT}: Graph-text joint representation learning for text generation
  from knowledge graphs}.
\newblock In \emph{Findings of the Association for Computational Linguistics:
  ACL-IJCNLP 2021}, pages 2526--2538, Online. Association for Computational
  Linguistics.

\bibitem[{Lango and Dusek(2023)}]{lango-dusek-2023-critic}
Mateusz Lango and Ondrej Dusek. 2023.
\newblock \href {https://doi.org/10.18653/v1/2023.emnlp-main.172}
  {Critic-driven decoding for mitigating hallucinations in data-to-text
  generation}.
\newblock In \emph{Proceedings of the 2023 Conference on Empirical Methods in
  Natural Language Processing}, pages 2853--2862, Singapore. Association for
  Computational Linguistics.

\bibitem[{Lavoie and Rainbow(1997)}]{lavoie-rainbow-1997-fast}
Benoit Lavoie and Owen Rainbow. 1997.
\newblock \href {https://doi.org/10.3115/974557.974596} {A fast and portable
  realizer for text generation systems}.
\newblock In \emph{Fifth Conference on Applied Natural Language Processing},
  pages 265--268, Washington, DC, USA. Association for Computational
  Linguistics.

\bibitem[{Lewis et~al.(2020)Lewis, Liu, Goyal, Ghazvininejad, Mohamed, Levy,
  Stoyanov, and Zettlemoyer}]{lewis-etal-2020-bart}
Mike Lewis, Yinhan Liu, Naman Goyal, Marjan Ghazvininejad, Abdelrahman Mohamed,
  Omer Levy, Veselin Stoyanov, and Luke Zettlemoyer. 2020.
\newblock \href {https://doi.org/10.18653/v1/2020.acl-main.703} {{BART}:
  Denoising sequence-to-sequence pre-training for natural language generation,
  translation, and comprehension}.
\newblock In \emph{Proceedings of the 58th Annual Meeting of the Association
  for Computational Linguistics}, pages 7871--7880, Online. Association for
  Computational Linguistics.

\bibitem[{{Llama Team}(2024)}]{llama3_2024}
{Llama Team}. 2024.
\newblock \href {https://doi.org/10.48550/arXiv.2407.21783} {The {Llama} 3
  {Herd} of {Models}}.
\newblock ArXiv:2407.21783 [cs].

\bibitem[{Papineni et~al.(2002)Papineni, Roukos, Ward, and jing
  Zhu}]{Papineni02bleu:a}
Kishore Papineni, Salim Roukos, Todd Ward, and Wei jing Zhu. 2002.
\newblock \href {https://www.aclweb.org/anthology/P02-1040} {{BLEU}: a method
  for automatic evaluation of machine translation}.
\newblock In \emph{Proceedings of the 40th annual meeting of the Association
  for Computational Linguistics}, pages 311--318, Philadelphia, PA, USA.

\bibitem[{Rebuffel et~al.(2022)Rebuffel, Roberti, Soulier, Scoutheeten,
  Cancelliere, and Gallinari}]{rebuffel_controlling_2022}
Clement Rebuffel, Marco Roberti, Laure Soulier, Geoffrey Scoutheeten, Rossella
  Cancelliere, and Patrick Gallinari. 2022.
\newblock \href {https://doi.org/10.1007/s10618-021-00801-4} {Controlling
  hallucinations at word level in data-to-text generation}.
\newblock \emph{Data Mining and Knowledge Discovery}, 36(1):318--354.

\bibitem[{Rozière et~al.(2023)Rozière, Gehring, Gloeckle, Sootla, Gat, Tan,
  Adi, Liu, Sauvestre, Remez, Rapin, Kozhevnikov, Evtimov, Bitton, Bhatt,
  Ferrer, Grattafiori, Xiong, Défossez, Copet, Azhar, Touvron, Martin,
  Usunier, Scialom, and Synnaeve}]{roziere_code_2024}
Baptiste Rozière, Jonas Gehring, Fabian Gloeckle, Sten Sootla, Itai Gat,
  Xiaoqing~Ellen Tan, Yossi Adi, Jingyu Liu, Romain Sauvestre, Tal Remez,
  Jérémy Rapin, Artyom Kozhevnikov, Ivan Evtimov, Joanna Bitton, Manish
  Bhatt, Cristian~Canton Ferrer, Aaron Grattafiori, Wenhan Xiong, Alexandre
  Défossez, Jade Copet, Faisal Azhar, Hugo Touvron, Louis Martin, Nicolas
  Usunier, Thomas Scialom, and Gabriel Synnaeve. 2023.
\newblock \href {https://doi.org/10.48550/arXiv.2308.12950} {Code {Llama}:
  {Open} {Foundation} {Models} for {Code}}.
\newblock ArXiv:2308.12950 [cs].

\bibitem[{Sellam et~al.(2020)Sellam, Das, and Parikh}]{bleurt}
Thibault Sellam, Dipanjan Das, and Ankur~P. Parikh. 2020.
\newblock \href {https://aclanthology.org/2020.acl-main.704/} {{BLEURT}:
  {Learning} {Robust} {Metrics} for {Text} {Generation}}.
\newblock In \emph{Proceedings of the 58th {Annual} {Meeting} of the
  {Association} for {Computational} {Linguistics}}, pages 7881--7892, Online.

\bibitem[{Touvron et~al.(2023)Touvron, Lavril, Izacard, Martinet, Lachaux,
  Lacroix, Rozière, Goyal, Hambro, Azhar, Rodriguez, Joulin, Grave, and
  Lample}]{touvron2023llama}
Hugo Touvron, Thibaut Lavril, Gautier Izacard, Xavier Martinet, Marie-Anne
  Lachaux, Timothée Lacroix, Baptiste Rozière, Naman Goyal, Eric Hambro,
  Faisal Azhar, Aurelien Rodriguez, Armand Joulin, Edouard Grave, and Guillaume
  Lample. 2023.
\newblock \href {http://arxiv.org/abs/2302.13971} {Llama: Open and efficient
  foundation language models}.

\bibitem[{White and Baldridge(2003)}]{white-baldridge-2003-adapting}
Michael White and Jason Baldridge. 2003.
\newblock \href {https://www.aclweb.org/anthology/W03-2316} {Adapting chart
  realization to {CCG}}.
\newblock In \emph{Proceedings of the 9th {E}uropean Workshop on Natural
  Language Generation ({ENLG}-2003) at {EACL} 2003}.

\bibitem[{Wolf et~al.(2020)Wolf, Debut, Sanh, Chaumond, Delangue, Moi, Cistac,
  Rault, Louf, Funtowicz, Davison, Shleifer, von Platen, Ma, Jernite, Plu, Xu,
  Le~Scao, Gugger, Drame, Lhoest, and Rush}]{wolf-etal-2020-transformers}
Thomas Wolf, Lysandre Debut, Victor Sanh, Julien Chaumond, Clement Delangue,
  Anthony Moi, Pierric Cistac, Tim Rault, Remi Louf, Morgan Funtowicz, Joe
  Davison, Sam Shleifer, Patrick von Platen, Clara Ma, Yacine Jernite, Julien
  Plu, Canwen Xu, Teven Le~Scao, Sylvain Gugger, Mariama Drame, Quentin Lhoest,
  and Alexander Rush. 2020.
\newblock \href {https://doi.org/10.18653/v1/2020.emnlp-demos.6} {Transformers:
  State-of-the-art natural language processing}.
\newblock In \emph{Proceedings of the 2020 Conference on Empirical Methods in
  Natural Language Processing: System Demonstrations}, pages 38--45, Online.
  Association for Computational Linguistics.

\end{thebibliography}

\appendix
\section{Prompts}
\label{app:prompts}

In Figures~\ref{fig:prompt1},~\ref{fig:prompt2} and~\ref{fig:prompt3}, we provide templates of prompts used in our approach for training a rule-based system. 

\begin{figure*}[tp]
\begin{lstlisting}
Complete Python code to convert given facts (triples) into a factual textual description (output). 
Write only a fragment of Python code that will replace the comment in the snippet below and nothing else. Do not include code that I have already written. triples is a list of tuples where each tuple is (subj, relation, obj).
Your code should be included inside this template:

triples = {triples}
relations = [triple.pred for triple in triples]
if (relations == {relations}):
     // your code to generate output
    output = ...
    print(output)

The output should be "{output}". The code should work even if the values of subj and obj in the triples are different, but the relations (pred) at the input of the program will always be the same and in the same order. Wrap any code in <code></code> tags.

\end{lstlisting}
\caption{Prompt used to generate rules in our approach.}
\label{fig:prompt1}
\end{figure*}

\begin{figure*}[tp]
\begin{lstlisting}
The desired output is: "{}"
but your code yields: "{}"
Could you produce code that returns the correct output? Remember to wrap the code in <code></code> tags.

\end{lstlisting}
\caption{Prompt used to inquire for rule edits in our approach.}
\label{fig:prompt2}
\end{figure*}

\begin{figure*}[tp]
\begin{lstlisting}
Your task is to create a sample for data-to-text dataset.
For a given set of relations generate a corresponding list of RDF triples and a text that describes them. Keep the same formating as in the example below. 
All the triples should be related (e.g. add information about already mentioned entities). 
The output text should ONLY describe the input triples and NOT add any extra information.

#### Example
relations: birth place, birth year, capital of
<sample>
in: (Mozart | birth place | Viena), (Mozart | birth year | 1756), (Vienna | capital of | Austria)
out: Mozart was born in 1756 in the capital of Austria, Vienna.
</sample>

#### Example
relations: {relations}
<sample>
in: {input}
out: {out}
</sample>
\end{lstlisting}
\caption{Prompt used to generate artificial training instances in our approach.}
\label{fig:prompt3}
\end{figure*}

In Figure~\ref{fig:promptd2t}, we show the prompt used for the zero-shot prompted LLM baseline to generate triple verbalizations directly.

All prompts are templates, with placeholders containing the specific data instances denoted by ``\texttt{\{name\}}", i.e.~they follow the Python string formatting convention.

\begin{figure*}[tp]
\begin{lstlisting}
You are given the following list of RDF triples.
{triples}
Write a plain text description of this data. Output only the text of the description.
\end{lstlisting}
\caption{Prompt for the zero-shot prompted LLM direct data-to-text generation baseline.}
\label{fig:promptd2t}
\end{figure*}

\section{Hyperparameters of BART fine-tuning}
\label{app:trainingdetails}

We used the BART-base model provided by the HuggingFace library.\footnote{\url{https://huggingface.co/facebook/bart-base}}
AdamW with learning rate $\eta = 2\cdot 10^{-5}$ and parameters $\beta=(0.9,0.997)$, $\epsilon = 10^{-9}$ was used as optimizer.\
Additionally, we applied  polynomial scheduler of $\eta$ with a warmup equal to 10\% of optimization steps.
The training was scheduled for 20 epochs with early stopping on validation loss (patience of 10 epochs).
We used batch size equal to 8 and label smoothing with $0.1$ smoothing factor.



\section{Examples of constructed rules}
\label{app:examplesrules}

In Figure~\ref{fig:examples}, we provide several examples of rules constructed by our approach. 

\begin{figure*}[tp]
\begin{subfigure}{\textwidth}
\begin{lstlisting}[language=Python]
subj = triples[0].subj
obj = triples[0].obj
relation = triples[0].pred
output = f"{subj} {relation} {obj}."
\end{lstlisting}
\caption{A simple rule to describe the ``is part of" relation.}
\end{subfigure}

\vspace{1cm}
\begin{subfigure}{\textwidth}
\begin{lstlisting}[language=Python]
subj = triples[0][0]
birth_date = next(obj for subj, pred, obj in triples if pred == 'birth date')
birth_place = next(obj for subj, pred, obj in triples if pred == 'birth place')
alma_mater = next(obj for subj, pred, obj in triples if pred == 'alma mater')
award = next(obj for subj, pred, obj in triples if pred == 'award')

output = f"{subj}, born on {birth_date} in {birth_place}, graduated from {alma_mater}, his alma mater. He won the prestigious {award}."
\end{lstlisting}
\caption{A rule for describing an input with the following set of relations: ``alma mater", ``award", ``birth date" and ``birth place".}
\end{subfigure}

\vspace{1cm}
\begin{subfigure}{\textwidth}
\begin{lstlisting}[language=Python]
subj = triples[0].subj
output = f"{triples[1].obj} is the {triples[1].pred} of {subj} located at {float(triples[2].obj):.0f} metres above sea level in {triples[0].obj}. The airport runway, named {triples[3].obj} has a length of {float(triples[4].obj):.0f}."
\end{lstlisting}
\caption{A rule for describing an input with the following set of relations: ``city served",
 ``operating organisation",
  ``elevation above the sea level",
  ``runway name" and
  ``runway length". Note the use of number formatting functions.}
\end{subfigure}

\vspace{1cm}
\begin{subfigure}{\textwidth}
\begin{lstlisting}[language=Python]
subj = triples[0].subj
industry_obj = [triple.obj for triple in triples if triple.pred == 'industry'][0]
product_obj = [triple.obj for triple in triples if triple.pred == 'product'][0]

if product_obj.lower() == 'world wide web':
    product_obj = 'web'

output = f"{subj} not only offers applications in the {industry_obj.lower()} industry, but also produces {product_obj} services."
\end{lstlisting}
\caption{A rule for describing an input with the following set of relations: ``industry", ``product". The rule overfitted to the training example related to web applications.}
\end{subfigure}

\caption{Examples of rules automatically constructed by our approach. Note that by default, the input is accessible to the rules via the ``\texttt{triples}'' list.}
\label{fig:examples}
\end{figure*}

\end{document}